\documentclass[letterpaper]{article} % DO NOT CHANGE THIS
\usepackage{aaai25}  % DO NOT CHANGE THIS
\usepackage{times}  % DO NOT CHANGE THIS
\usepackage{helvet}  % DO NOT CHANGE THIS
\usepackage{courier}  % DO NOT CHANGE THIS
\usepackage[hyphens]{url}  % DO NOT CHANGE THIS
\usepackage{graphicx} % DO NOT CHANGE THIS
\usepackage{natbib}  % DO NOT CHANGE THIS AND DO NOT ADD ANY OPTIONS TO IT
\usepackage{caption} % DO NOT CHANGE THIS AND DO NOT ADD ANY OPTIONS TO IT
\usepackage{algorithm}
\usepackage{algorithmic}
\usepackage{xcolor}
\newtheorem{theorem}{Theorem}
\newtheorem{cor}{Corollary}

\newtheorem{lemma}{Lemma}
\newtheorem{example}{Example}

\newtheorem{remark}{Remark}
\newtheorem{ass}{Assumption}
\newtheorem{definition}[cor]{Definition}
\newcommand{\Norm}[1]{\left\|#1\right\|}
\usepackage{makecell} 
\usepackage{tabularx}
\usepackage{booktabs}
\usepackage{subfigure}
\usepackage{threeparttable}
\usepackage{multirow}
\usepackage{makecell}
\usepackage{amssymb}
\usepackage{mathtools}
\def \E {\mathbb{E}}

\def \g {\mathbf{g}}

\def \x {\mathbf{x}}
\def \y {\mathbf{y}}

\def \F {\mathcal{F}}

\def \X {\mathcal{X}}

\usepackage{newfloat}
\usepackage{listings}
\DeclareCaptionStyle{ruled}{labelfont=normalfont,labelsep=colon,strut=off} % DO NOT CHANGE THIS
\floatstyle{ruled}
\newfloat{listing}{tb}{lst}{}
\floatname{listing}{Listing}
\title{Online Nonsubmodular Optimization with\\ Delayed Feedback in the Bandit Setting}
\author{
    Sifan Yang\textsuperscript{\rm 1,2}, \  
    Yuanyu Wan\textsuperscript{\rm 3,1,}\footnotemark[1], \
    Lijun Zhang\textsuperscript{\rm 1,3,}\thanks{  Corresponding author}
}
\affiliations {
\textsuperscript{\rm 1}National Key Laboratory for Novel Software Technology, Nanjing University, Nanjing, China \\
\textsuperscript{\rm 2}School of Artificial Intelligence, Nanjing University, Nanjing, China \\
\textsuperscript{\rm 3}State Key Laboratory of Blockchain and Data Security, Zhejiang University, Hangzhou, China\\
\{yangsf,zhanglj\}@lamda.nju.edu.cn, wanyy@zju.edu.cn
}

\begin{document}

\maketitle

\begin{abstract}
We investigate the online nonsubmodular optimization with delayed  feedback in the bandit setting, where the loss function is  $\alpha$-weakly DR-submodular and $\beta$-weakly DR-supermodular.  Previous work has established an  $(\alpha,\beta)$-regret bound of $\mathcal{O}(nd^{1/3}T^{2/3})$, where $n$ is the  dimensionality and $d$ is the maximum delay. However, its regret  bound relies on the maximum delay and is thus sensitive to irregular delays. Additionally,  it couples the effects of delays and bandit feedback as its bound is the product of the delay term and the $\mathcal{O}(nT^{2/3})$ regret bound in the bandit setting without delayed feedback.  In this paper, we develop two algorithms to address these limitations, respectively. Firstly,  we propose a novel method, namely DBGD-NF, which employs the one-point gradient estimator and utilizes all the available estimated gradients in each round to update the decision. It achieves a better $\mathcal{O}(n\bar{d}^{1/3}T^{2/3})$ regret bound, which is relevant to the average delay $\bar{d} = \frac{1}{T}\sum_{t=1}^T d_t\leq d$. Secondly,  we extend DBGD-NF by employing a blocking update mechanism to decouple the joint effect of the delays and bandit feedback, which enjoys an  $\mathcal{O}(n(T^{2/3} + \sqrt{dT}))$ regret bound.     When $d = \mathcal{O}(T^{1/3})$, our regret bound matches the  $\mathcal{O}(nT^{2/3})$  bound  in the bandit setting without delayed feedback. Compared to our first $\mathcal{O}(n\bar{d}^{1/3}T^{2/3})$ bound, it is more advantageous when the maximum delay $d = o(\bar{d}^{2/3}T^{1/3})$.  Finally, we conduct experiments on structured sparse learning to demonstrate the superiority of our methods.

\end{abstract}

% Uncomment the following to link to your code, datasets, an extended version or similar.
%
% \begin{links}
%     \link{Code}{https://aaai.org/example/code}
%     \link{Datasets}{https://aaai.org/example/datasets}
%     \link{Extended version}{https://aaai.org/example/extended-version}
% \end{links}

\section{Introduction}
Online learning is a powerful framework that has been used to model  various sequential prediction  problems \cite{shalev2012online}. 
It can address scenarios in which decisions are made from a small set  \cite{hazan2012online}, a continuous space \cite{hazan2016introduction}, or a combinatorial feasible domain \cite{cesa2012combinatorial}. 
In this paper, we study the online nonsubmodular optimization \cite{lin2022online}, an emerging branch of the online learning, which appears in many machine learning tasks like structured sparse learning \cite{el2015totally}, Bayesian optimization \cite{gonzalez2016batch}, and column subset selection \cite{sviridenko2017optimal}, etc.   Similar to the classical online convex optimization (OCO) \cite{zinkevich2003online}, it is typically formulated as a game between an online player and an adversary. In each round $t \in [T]$, the player begins by selecting a subset decision $S_t \subseteq [n]$. When the player submits its decision $S_t$, the adversary  chooses a nonsubmodular loss function $f_t(\cdot): 2^{[n]} \mapsto \mathbb{R}$ and then the player suffers a loss $f_t(S_t)$, where  $f_t(\cdot)=\bar{f}_t\left(\cdot\right)- \underline{f}_t\left(\cdot\right)$, $\bar{f}_t\left(\cdot\right)$  is $\alpha$-weakly diminishing return submodular (DR-submodular),  $\underline{f}_t\left(\cdot\right)$ is $\beta$-weakly diminishing return supermodular (DR-supermodular) and  $2^{[n]}$ represents all the subset of $[n]$.\footnote{The definition of the $\alpha$-weakly DR-submodular and $\beta$-weakly DR-supermodular can be found in Definition \ref{definition2}.}

The player aims to minimize the cumulative loss over $T$ rounds,  equivalently  minimizing the regret:
 \begin{equation}\label{regret}
     \textbf{Reg}(T) \triangleq \sum_{t=1}^T f_t(S_t) - \min_{S \subseteq [n]} \sum_{t=1}^T f_t(S),
 \end{equation}
 which compares the excess loss suffered by the player with that of the best decision chosen in hindsight.  As pointed out by \citet{el2020optimal}, the optimization problem $\min_{S \subseteq [n]} \sum_{t=1}^T f_t(S)$ is NP-hard, thus it is impossible to find the optimal decision in polynomial time. For this reason,  we  follow the previous work  \cite{lin2022online} and  apply the $(\alpha,\beta)$-regret to measure the performance of the online player,     which is defined as 
\begin{equation}\label{newregret}
    \textbf{Reg}_{\alpha,\beta}(T) \triangleq \sum_{t=1}^Tf_t(S_t) - \left(\frac{1}{\alpha} \bar{f}_t\left(S^{\star}\right)-\beta \underline{f}_t\left(S^{\star}\right)\right),
\end{equation}
where  $S^* = \arg \min_{S \subseteq [n]} \sum_{t=1}^T f_t(S)$ and  $(\alpha, \beta)$ are the approximation factors achieved by a certain offline algorithm.    \citet{lin2022online} are the first to investigate the online nonsubmodular optimization and develop a method that achieves an $(\alpha,\beta)$-regret bound of $\mathcal{O}(\sqrt{nT})$, building on the Lovász extension \cite{lovasz1983submodular} and the convex relaxation model \cite{el2020optimal}, where $n$ is the dimensionality.

\begin{table}[t]
\begin{center}
\begin{small}
\begin{tabular}{cccc}
\toprule
Setting & Method & $(\alpha,\beta)$-regret bound    \\
\midrule
bandit & DBAGD \cite{lin2022online}& $\mathcal{O}\left(nd^{1/3}T^{2/3}\right)$   \\
\midrule
bandit & BDGD-NF (\textbf{Theorem} \ref{thm:2})& $\mathcal{O}\left(n\Bar{d}^{1/3}T^{2/3}\right)$ \\
\midrule
bandit & BDBGD-NF (\textbf{Theorem} \ref{thm:1}) & $\mathcal{O}\left(n(T^{2/3}+\sqrt{dT})\right)$  \\
\midrule
full & DOAGD \cite{lin2022online}& $\mathcal{O}\left(\sqrt{ndT}\right)$   \\
\midrule
full & DOGD-NF (\textbf{Theorem} \ref{thm:0}) & $\mathcal{O}\left(\sqrt{n\Bar{d}T}\right)$  \\
\bottomrule
\end{tabular}
\end{small}
\end{center}
\caption{Summary of results for online nonsubmodular optimization under different settings, where $n$ is the dimensionality, $d$ is the maximum delay and $\Bar{d} = \frac{1}{T}\sum_{t=1}^T d_t$ is the average delay over $T$ rounds. For simplicity, we use the abbreviations:  full $\rightarrow$  full-information setting with delayed feedback, bandit $\rightarrow$  bandit setting with delayed feedback.}\label{tab:1}
\end{table}
% However,  in some practical applications,  the online player commonly does not receive any additional information (such as gradient) about the loss function $f_t(\cdot)$ beyond the value $f_t(S_t)$.  For example, in online advertising,  a recommendation algorithm suggests the advertisements to users, and the decision involves choosing ad delivery strategies. Advertisers adjust their strategies based on the interactions of users, such as advertisement clicks, while the response is just a value, and there is often a delay in receiving the response \cite{he2014practical,heliou2020gradient,wan2022online}.
% To address this issue, we also investigate online nonsubmodular optimization with delayed and bandit feedback in this paper.

% However, in many real world scenarios, there may be a potential delay between the query of the player and the response. For example, in online advertising, the decision is the ad delivery strategies for users.  Advertisers usually adjust their strategies based on user interactions, such as advertisement clicks, while there is often a delay in receiving the feedback.
% ~\citep{mcmahan2013ad,joulani2013online,he2014practical,wan2022online}.

 In lots of real-world scenarios, there may be a potential delay between the query of the player and the corresponding response \cite{quanrud2015online,wan2022projection,wan2023non}. 
% While the formulation for online nonsubmodular optimization is well-defined, practical applications often involve significantly greater complexity, including challenges such as non-stationary environments \cite{wan2021projection,wan2023improved,wangonline}, delayed feedback \cite{wan2022online,wan2022projection,wan2023non}, and partial feedback (bandit feedback) \cite{flaxman2005online,saha2011improved,AAAI:2015:Zhang}, etc.
 To address the delayed scenario, \citet{lin2022online} also explore  the problem of online nonsubmodular optimization with delayed feedback, where the online player incurs an  arbitrary delay $d_t \geq 1$ in receiving the response. To handle the delayed feedback, they utilize the pooling strategy \cite{heliou2020gradient} to propose delay online approximate gradient descent (DOAGD), which enjoys an $\mathcal{O}(\sqrt{ndT})$  regret bound, where    $d$ is the maximum delay. Moreover, they consider a more challenging setting,  bandit feedback, in which the online player does not receive any additional information about the loss function $f_t(\cdot)$ (e.g., its gradient) beyond the value $f_t(S_t)$, and develop    delay online approximate gradient descent (DBAGD), achieving an $\mathcal{O}(nd^{1/3}T^{2/3})$  regret bound. However, their result for  bandit setting with delayed feedback, summarized in Table \ref{tab:1}, has two limitations. Firstly, it relies on the maximum delay $d$,  which renders it  sensitive to  irregular delays.  Secondly, it is the product of the delay   term and the $\mathcal{O}(nT^{2/3})$ regret bound in the non-delayed bandit setting. This arises from DBAGD coupling the effects of the delays and bandit feedback, resulting in a discontented regret bound.

To overcome these limitations, we revisit the online nonsubmodular optimization with delayed feedback in the bandit setting. Specifically, we first develop a delayed algorithm to establish a regret bound that is relevant to the average delay. Our proposed method, named    delayed bandit gradient descent for nonsubmodular function~(DBGD-NF),  achieves an enhanced $\mathcal{O}\left(n\Bar{d}^{1/3}T^{2/3}\right)$ regret bound, where $\bar{d} = \frac{1}{T}\sum_{t=1}^{T}d_t \leq d$ represents the average delay. The primary idea is to employ the one-point gradient estimator \cite{hazan2012online,lin2022online} and use all available estimated gradients to update the decision in each round,  instead of utilizing the oldest one like  DBAGD.  
Furthermore, as a by-product, if the online player has access to the gradient of the loss function, we  can substitute the estimated gradient in DBGD-NF with the true gradient. Our algorithm, referred to as delayed online gradient descent for nonsubmodular function (DOGD-NF),  enjoys a better $\mathcal{O}(\sqrt{n\Bar{d}T})$ regret bound for the full-information setting with delayed feedback.

In our pursuit of decoupling the joint effect of delayed feedback and the gradient estimator, we develop the blocking delayed bandit gradient descent for nonsubmodular function~(BDBGD-NF).   Drawing inspiration from  \citet{wan2024improved}, we adopt the blocking update mechanism \cite{zhang2019online,garber2020improved,AAAI:2023:Wang,AAAI:2024:Wang} with  DBGD-NF. Particularly, we divide the total $T$ rounds into several blocks of size $K$ and update the decision at the end of each block using the estimated gradients from the blocks where all gradients are available. 
By setting a appropriate block size $K$, we can reduce the variance of the one-point gradient estimator. Leveraging this technique, BDBGD-NF  achieves a superior  $\mathcal{O}(n(T^{2/3}+\sqrt{dT}))$ regret bound. When the algorithm faces small delays, i.e., maximum delay $d = \mathcal{O}(T^{1/3})$, this regret bound matches  the existing $\mathcal{O}(nT^{2/3})$ regret bound in the non-delayed setting \cite{lin2022online}, benefiting from the blocking update mechanism.

  On the other hand, when the impact of the delayed feedback $d$ is substantial, i.e., $d = \Omega (T^{1/3})$, our regret bound is on the same order as the $\mathcal{O}(\sqrt{n\Bar{d}T})$ bound we establish for the full-information setting in terms of $d$ and $T$ under the worst case, where $\bar{d} = \Theta(d)$. Moreover, it is better than our first $\mathcal{O}\left(n\Bar{d}^{1/3}T^{2/3}\right)$ bound when the maximum delay $d = o(\Bar{d}^{2/3}T^{1/3})$. Notably, compare to \citet{wan2024improved}, BDBGD-NF is specifically designed for online nonsubmodular optimization, making it more challenging to analyze.   Finally, we compare our  algorithms with the state-of-the-art (SOTA) methods through numerical experiments   to demonstrate the robustness and  effectiveness in handling delayed and bandit feedback effects.

To summarize, this paper makes the following contributions to online nonsubmodular optimization with  delayed feedback:
\begin{itemize}
\item We propose two algorithms for online nonsubmodular optimization with delayed feedback to derive the regret bounds that are relevant to the average delay. Our methods reduce the regret bounds to $\mathcal{O}(\sqrt{n\Bar{d}T})$ and $\mathcal{O}\left(n\Bar{d}^{1/3}T^{2/3}\right)$ for full-information and bandit feedback settings, respectively. 

\item  To decouple the joint effect of the delayed and bandit feedback,  we develop a novel algorithm by utilizing a blocking update technique, which enjoys an   $\mathcal{O}(n(T^{2/3}+\sqrt{dT}))$ regret bound.

\end{itemize}

\section{Related Work}
In this section, we briefly introduce the related work of  submodular optimization and  nonsubmodular optimization.

\subsection{Submodular Optimization}
Submodular optimization has garnered increasing interest  in  various practical applications, such as sparse reconstruction \cite{bach2010structured,das2012selecting,liao2023improved}, graph inference \cite{rodriguez2012submodular,defazio2012convex}, video analysis \cite{zheng2014submodular}, object detection \cite{song2014learning}, etc. The primary property of a submodular function is diminishing returns, meaning that adding an element to a larger set provides less incremental gain than adding it to a smaller set.
Early research \cite{badanidiyuru2014fast,krause2014submodular} on submodular optimization mainly focuses on the offline setting, which may be unsuitable for sequential decision-making problems. To overcome this issue,  \citet{hazan2012online} investigate the online submodular optimization, where the player chooses a subset from a ground set of $n$ elements in round $t$, and then observe a submodular loss function. Based on the Lovász extension \cite{lovasz1983submodular}, they extend online gradient descent (OGD) \cite{zinkevich2003online} and bandit gradient descent (BGD) \cite{flaxman2005online} into submodular optimization in both full-information and bandit feedback settings, establishing $\mathcal{O}(\sqrt{nT})$ and $\mathcal{O}(nT^{2/3})$ regret bounds, respectively,  where $n$ is the dimensionality.

\subsection{Nonsubmodular Optimization}

Although submodularity is a natural assumption,  the objective function is not always exactly submodular in various applications, such as structured sparse learning \cite{el2015totally}, Bayesian optimization \cite{gonzalez2016batch}, and column subset selection \cite{sviridenko2017optimal}, etc. Instead, it satisfies a weaker version of the DR property, like  $\alpha$-weakly DR-submodular  and $\beta$-weakly DR-supermodular \cite{lehmann2006combinatorial}. \citet{el2020optimal}  provide the first approximation guarantee for  nonsubmodular minimization by developing an approximate projected subgradient method. Nevertheless, they only  focus on the offline  setting.

 \citet{lin2022online} pioneer the study of online nonsubmodular optimization and introduce the $(\alpha,\beta)$-regret to measure the performance of the online player, which is defined in (\ref{newregret}). Based on the Lovász extension \cite{lovasz1983submodular} and the convex relaxation model \cite{el2020optimal}, they propose  online approximate gradient descent (OAGD), which obtains an $\mathcal{O}(\sqrt{nT})$  regret bound. Particularly, it uses the subgradient of the convex relaxation function to perform a gradient descent step over the Lovász extension domain. Then it samples the decision $S_t$ from a certain distribution  over all possible sets  at  round $t$. Moreover, they also consider the full-information and bandit settings with  delay feedback, which are more complex to analyze. To handle the delay feedback, they further extend OAGD by adopting a pooling strategy \cite{heliou2020gradient}. Their method, namely DOAGD,   which keeps a pool to store all the available delayed information and utilizes the oldest received but not utilized  gradient to update the decision in each round,  achieves an $\mathcal{O}(\sqrt{ndT})$ regret bound, where $d$ is the maximum delay. Nevertheless, it  depends on the maximum delay, and thus is  sensitive to irregular delays.   In the bandit setting, since the player can only observe the loss value of its decision $S_t$, they employ the importance weighting technique to estimate the gradient. In particular, $S_t$ is chosen from a distribution that is related to the decision  with probability $1-\mu$ ($0<\mu<1$) and  a random distribution with probability $\mu$ for exploration, ensuring the variance of the gradient is upper bounded by $\mathcal{O}(n^2/\mu)$.  While they establish an $\mathcal{O}(nd^{1/3}T^{2/3})$ regret bound, their method couples the effects of the delays and gradient estimator.

 \section{Preliminary}
In this section, we will provide  essential definitions and basic setup for optimization of the nonsubmodular functions.

\subsection{Definitions and Assumptions}
\begin{definition}
    For any function $f(\cdot) : 2^{[n]} \mapsto \mathbb{R}$, we define $f(i \mid S) = f(\{i\} \bigcup S) - f(S)$ to denote the marginal gain of adding an element $i$ to $S$. Moreover, $f(\cdot)$ is normalized if and only if $f(\emptyset)=0$ and nondecreasing if and only if $f(A)\leq f(B)$ for any $A \subseteq B$. $\Pi_{[0,1]^n}$ is the projection onto the domain $[0,1]^n$, which can be efficiently implemented as a simple clipping operation.
\end{definition}

\begin{definition}\label{definition2}
    A function $f(\cdot) : 2^{[n]} \mapsto \mathbb{R}$ is $\alpha$-weakly DR-submodular with $\alpha > 0$ if 
    \begin{equation*}
        f(i \mid A) \geq \alpha f(i \mid B), \quad  \text{for all }  A \subseteq B, i \in[n] \backslash B.
    \end{equation*}
If this inequality holds when $\alpha=1$, $f(\cdot)$ is submodular.

Similarly, $f(\cdot) : 2^{[n]} \mapsto \mathbb{R}$ is $\beta$-weakly DR--supermodular with $\beta > 0$ if 
$$f(i \mid B) \geq \beta f(i \mid A), \quad  \text{for all }  A \subseteq B, i \in[n] \backslash B.$$
$f(\cdot)$ is supermodular when $\beta=1$.

We define that $f(\cdot)$ is $(\alpha,\beta)$-weakly DR-modular if both of the above inequalities hold simultaneously.
\end{definition}

Building on the above definitions, we formulate the problem of minimizing structured nonsubmodular functions \cite{el2020optimal,lin2022online}:
\begin{equation}\label{nonsub}
    f(S):=\bar{f}(S)-\underline{f}(S),
\end{equation}
where $S \subseteq [n]$. Afterwards,  we introduce two common assumptions in the online nonsubmodular optimization.

\begin{ass}\label{ass:1}
    All the nonsubmodular functions $f(\cdot)$ defined in (\ref{nonsub}) satisfy $\bar{f}(S)+\underline{f}(S) \leq L$ for all $S \subseteq [n]$.
\end{ass}

\begin{ass}\label{ass:2}
 $\bar{f}(\cdot)$ and $\underline{f}(\cdot)$ defined in (\ref{nonsub}) are normalized and non-decreasing. $\bar{f}(\cdot)$ is $\alpha$-weakly DR-submodular and $\underline{f}(\cdot)$ is $\beta$-weakly DR-supermodular.
\end{ass}

Then we give an application of the online nonsubmodular optimization to provide practical insights of these assumptions, which is also used in our later experiments.

\textbf{Structured sparse learning.} This problem aims to learn a sparse parameter vector whose support satisfies a specific structure, such as group-sparsity, clustering, tree-structure, or diversity \cite{kyrillidis2015structured}. 
It is typically formulated  as $\min _{\x \in \mathbb{R}^{n}} \ell(\x)+\gamma F(\operatorname{supp}(\x))$, where $\ell(\cdot) : \mathbb{R}^{n} \mapsto \mathbb{R}$ is the loss function, $F(\cdot): 2^{[n]} \mapsto \mathbb{R}$ is  a set function that imposes restrictions on the support set and $\gamma $ is a trade-off parameter. Previous approaches \cite{bach2010structured,el2015totally}  often replace the discrete regularizer $F(\operatorname{supp}(\x))$ with its closest convex relaxation, which is computationally tractable only when $F(\cdot)$ is submodular. \citet{el2020optimal} introduce an alternative formulation by using a nonsubmodular regularizer, which is better in practice, defined as 
 \begin{equation}\label{exper}
     \min_{S \subseteq[n]} H(S) = \gamma F(S)-G(S),
 \end{equation}
where $G(S) = \ell(0) - \min_{\operatorname{supp}(\x) \subseteq S} \ell(\x)$ is a normalized non-decreasing set function. \citet{el2020optimal} have pointed out that when $\ell(\cdot)$ is smooth, strongly convex and is generated from random data, $G(\cdot)$ is weakly DR-modular. Moreover, if $F(\cdot)$ is $\alpha$-weakly DR-submodular, the equation (\ref{exper}) can be transformed into (\ref{nonsub}) so that we can handle it directly. For example,  $F(\cdot)$ is often chosen as the range cost function \cite{bach2010structured} ($\alpha = \frac{1}{n-1}$ and $n$ is the dimensionality),  applied in the time-series and cancer diagnosis \cite{rapaport2008classification},  and  the cost function  ($\alpha = \frac{1+a}{1+b-a}$ and $a,b$ are cost parameters), applied in the healthcare \cite{sakaue2019greedy}.

Since nonsubmodular functions are defined over the discrete domain, determining their minimum values is a challenging task. Therefore, we introduce the Lovász extension \cite{lovasz1983submodular} to transform a function $f(\cdot)$ defined over a discrete domain $[n]$ to a new function $f_L(\cdot)$ over the unit hypercube $[0,1]^n$. The extended function $f_L(\cdot)$ is convex if and only if $f(\cdot)$ is submodular. For nonsubmodular functions, we can exploit the convex relation \cite{el2020optimal}, enabling the
use of convex optimization algorithms on the transformed function.

\subsection{Lovász Extension and Convex Relaxation}
Lovász extension ensures that the minima of the function over the domain $[0,1]^n$  also recover the minima of the original function $f(\cdot)$. In this way, we can reduce the complex optimization task over domain $[n]$ to a simpler convex optimization problem. To clarify this reduction process, we start with some necessary definitions.

\begin{definition}
    A max chain of subsets of $[n]$ is a collection of sets $\{A_0,...,A_n\}$, and   $ \emptyset = A_0 \subseteq A_1 \subseteq ... \subseteq A_n = [n]$.
\end{definition}

For any $\x \in [0,1]^n = \X$, we introduce a unique associated permutation  $\pi : [n] \mapsto [n]$ such that $\pi(i) = j$,  meaning that $\x_j$ is the $i$-th largest number in $\x$. Notably, we have $1\geq \x_{\pi(1)} \geq ... \geq \x_{\pi(n)} \geq  0$ and  let $\x_{\pi(0)} = 1 ,\x_{\pi(n+1)}=0$ for simplicity.  If we set $A_i = \{\pi(1),...,\pi(i)\}$ for all $i \in [n]$ and $A_0 =\emptyset$, the vector $\x$ can be expressed as a convex combination, i.e., $\x = \sum_{i=0}^{n} \lambda_{i} \chi(A_{i})$, where $\lambda_i = \x_{\pi(i)} - \x_{\pi(i+1)}$ and $\sum_{i=0}^n \lambda_i =1, \lambda_i \in [0,1]$ \cite{hazan2012online}. For any set $S \subseteq [n]$, $\chi(\cdot): 2^{[n]} \mapsto \{0,1\}^{n}$ is an indicator function $\chi(S)_i=1 $ for all $ i \in S$ and $\chi(S)_i=0 $ for all $ i \notin S$.  Next, we give the definition of the Lovász extension.

\begin{definition}
    For any submodular function $f(\cdot)$, its Lovász extension $f_L(\cdot) : \X = [0,1]^n \mapsto \mathbb{R}$ is defined as  $f_L(\x) = \sum_{i=0}^n (\x_{\pi(i)} - \x_{\pi(i+1)}) f(A_i) = \sum_{i=0}^n\x_{\pi(i)} f(\pi(i)\mid A_{i-1}) .$
\end{definition}

It is not hard to verify that $f_L(\chi(S)) = f(S)$ for any $S \subseteq [n]$. 
% Furthermore, since the Lovász Extension is convex, its minimum value on the domain $[0,1]^n$ will be  at a vertex, which corresponds to a set in the original discrete domain. 
Therefore, minimizing the Lovász extension is equivalent to minimizing the original submodular function over all possible sets. Moreover,   \citet{edmonds2003submodular} has pointed out that the subgradient $\g$ of $f_L(\x)$ can be computed by 
\begin{equation}\label{subg}
    \g_{\pi(i)}=f\left(A_{i}\right)-f\left(A_{i-1}\right)
\text{ for all }i \in[n].
\end{equation}

However, when $f(\cdot)$ is not submodular, many properties break down. For example, $f_L(\cdot)$ is non-convex, which is harder to analyze. To tackle the nonsubmodular functions, we  adopt the convex closure $f_C(\cdot)$, which is defined as:
\begin{definition}\label{def6}
    The convex closure $f_C(\cdot): [0,1]^n \mapsto \mathbb{R}$ for a nonsubmodular function $f(\cdot)$ is the point-wise largest convex function which always lower bounds $f(\cdot)$. Additionally, $f_C(\cdot)$ is the tightest convex extension of $f(\cdot)$ and $\min _{S \subseteq[n]} f(S)=\min _{x \in [0,1]^n} f_{C}(\x)$.
\end{definition}

Definition \ref{def6} gives us a simpler way to analyze the nonsubmodular function.
Unfortunately, it is NP-hard to evaluate and optimize $f_C(\cdot)$ \cite{vondrak2007submodularity}. Nevertheless, we can utilize  the proposition 3.1 in \citet{lin2022online} to derive the approximation:

\begin{lemma}\label{lem:0}
 Assuming $f(\cdot)$ satisfies Assumption \ref{ass:2} and $\g$ is calculated according to (\ref{subg}) for all $A\subseteq [n] $ and $ \x \in \X$, we have the following guarantees
\begin{align*}
    f_{L}(\x)&=\left\langle \g, \x\right\rangle \geq f_{C}(\x),\\
    \sum_{i \in A} \g_{i} &\leq \frac{1}{\alpha} \bar{f}(A)-\beta \underline{f}(A),\\
    f_C(\x) \leq  f_L(\x) = & \left\langle \g, \x\right\rangle  \leq \frac{1}{\alpha} \bar{f}_{C}(\x) - \beta\underline{f}_{C}(\x).
\end{align*}
\end{lemma}

\begin{remark}\normalfont
    Lemma \ref{lem:0} demonstrates how Lovász extension $f_{L}(\x)$ approximates the convex closure $f_{C}(\x)$ so that the subgradient of $f_{L}(\x)$ can serve as the approximate subgradient for $f_C(\x)$ \cite{el2020optimal}, which plays an important role in our analysis. 
    % Moreover, it ensures that $(\alpha,\beta)$-regret is a reasonable metric. 
\end{remark}
\subsection{Problem Setup}

We consider the online nonsubmodular optimization with delayed feedback in the bandit setting, where the  loss function $f_t(\cdot)$ is defined in (\ref{nonsub}), and satisfies Assumption \ref{ass:1} and Assumption \ref{ass:2}. In each round $t$, the player makes a decision $S_t \subseteq [n]$ and then triggers a delay $d_t$ when receiving the loss value. The response will arrive at round $t+d_t-1$ and the player receives $\{f_k(S_k)| k \in \F_t\}$,  where $\mathcal{F}_{t}=\left\{k \mid k+d_{k}-1=t\right\}$
represents the index set of received loss values in round $t$.   To measure the performance of the online player,  we  follow the previous work  \cite{lin2022online} and apply the $(\alpha,\beta)$-regret  defined in (\ref{newregret}).   It compares the loss of the player's decisions to the result returned by an offline algorithm that approximately solves the optimization problem  $\min_{S \subseteq [n]} \sum_{t=1}^T f_t(S)$ in polynomial time, which is different from the vanilla regret \cite{zinkevich2003online}.  
% In particular, $(\alpha, \beta)$ are bounds on the quality of a result returned by some algorithm compared to the best solution.
% As pointed out by \citet{el2020optimal}, such approximation is optimal.

\section{Main Results}

In this section, we first develop a delayed algorithm for bandit setting to establish a regret bound that is relevant to the average delay. As a by-product, we demonstrate that it can be slightly adjusted into the full-information setting to obtain a better regret bound. Finally,  we present our blocking method to decouple the joint effect of the delays and bandit feedback, further enhancing the regret bound.

\begin{algorithm}[!t]
\caption{DBGD-NF}
	\label{alg:-}
	\begin{algorithmic}[1]
     \REQUIRE Learning rate $\eta$
     \STATE Initialize $\x_1 \in [0,1]^n$
     \FOR{$t=1$ to $T$}
     \STATE Let $1 \geq \x_{t,\pi_{(1)}} \geq \x_{t,\pi_{(2)}} \geq \dots \x_{t,\pi_{(n)}} \geq 0 $ be the sorted entries in decreasing order with $A_{t,i} =\{\pi_{(1)}, \dots, \pi_{(i)}\}$ for all $i \in [n]$ and $A_{t,0} = \emptyset$. Define $\x_{t,\pi_{(0)}}=1, \x_{t,\pi_{(n+1)}}=0$
     \STATE For $0 \leq i \leq n$, calculate $\lambda_{t,i} = \x_{t,\pi_{(i)}}-\x_{t,\pi_{(i+1)}}$
     \STATE Sample $S_t$ from the distribution $P \left(S_{t}=A_{t,i}\right) = (1-\mu)\lambda_{t,i} + \frac{\mu}{n+1}$
     \STATE Observe the loss $f_t(S_t)$ 
     \STATE Calculate $\hat{f}_{t,i}=\frac{\mathbf{1}\left(S_{t}=A_{t,i}\right)}{\left(1-\mu\right) \lambda_{t,i}+\frac{\mu}{n+1}} f_{t}\left(S_{t}\right)$
     \STATE Compute the estimated gradient $\hat{\g}$ $\hat{\g}_{t,\pi(i)}=\hat{f}_{t,i}-\hat{f}_{t,i-1}$ and incur a delay $d_t \geq 1$
     \STATE Receive the gradient set $\{\hat{\g}_k|k\in\F_t\}$
     \STATE Update  $ \x_{t+1} = \Pi_{[0,1]^n} \left[ \x_t - \sum_{k \in \F_t} \hat{\g}_k \right]$
     \ENDFOR
	\end{algorithmic}
\end{algorithm}

\subsection{Results Related to the Average Delay}

Existing literature \cite{lin2022online} on online nonsubmodular optimization with delayed feedback in the bandit setting adopts a pooling strategy \cite{heliou2020gradient} to handle the arbitrary delays, which only uses the oldest available information in a gradient pool. Since the gradient may be delayed by $d$ rounds, its regret bound relies on the maximum delay $d$. To  mitigate the effect of delayed feedback, motivated by previous work \cite{quanrud2015online}, we  utilize all the gradients received in round  $t$ to update the decision,  rather than the oldest one. In the bandit setting, the online player only has access to the loss value. To deal with this issue, we  employ the one-point estimator \cite{hazan2012online,lin2022online} to compute the unbiased gradient.  Our method, DBGD-NF, is detailed in Algorithm \ref{alg:-}. In each round $t$,   $S_t$ is sampled from the distribution
\begin{equation}\label{distribution}
    P \left(S_{t}=A_{t,i}\right) = (1-\mu)\lambda_{t,i} + \frac{\mu}{n+1},
\end{equation}
where $\lambda_{t,i} =  \x_{t,\pi_{(i)}}-\x_{t,\pi_{(i+1)}}$ and $\mu \in (0,1)$ is  the exploration probability.
Then we  utilize the one-point estimator  to derive the gradient; that is, we compute
\begin{equation}\label{le}
    \hat{f}_{t,i}=\frac{\mathbf{1}\left(S_{t}=A_{t,i}\right)}{\left(1-\mu\right) \lambda_{t,i}+\frac{\mu}{n+1}} f_{t}\left(S_{t}\right),
\end{equation}
 where $\mathbf{1}(\cdot)$ is an indicator function, and calculate the unbiased gradient
\begin{equation}\label{gradient}
    \hat{\g}_{t,\pi(i)}=\hat{f}_{t,i}-\hat{f}_{t,i-1}.
\end{equation}
Notably, we do not assume that the information is immediately available. Instead,  owing to incurring a delay $d_t$,  we only receive the information at the end of the round $t + d_t -1$, and we only present the calculation of the gradient for simplicity. Then we use all the available estimated gradients to update the decision
 \begin{equation}\label{update2}
    \x_{t+1} = \Pi_{[0,1]^n} \left[ \x_t - \sum_{k \in \F_t} \hat{\g}_k \right]
\end{equation}

Next, we establish the regret bound of Algorithm \ref{alg:-}.

\begin{theorem}\label{thm:2}
Under Assumption \ref{ass:1} and Assumption \ref{ass:2}, by setting $\mu = \frac{n\Bar{d}^{1/3}}{T^{1/3}}, \eta = \frac{1}{L\Bar{d}^{1/3}T^{2/3}}$, DBGD-NF ensures 
\begin{align*}
    \E\left[\textbf{Reg}_{\alpha,\beta}(T)\right] \leq \mathcal{O}\left(n\Bar{d}^{1/3}T^{2/3}\right),
\end{align*}
    where $n$ is the dimensionality and $\Bar{d} = \frac{1}{T} \sum_{t=1}^T d_t$ is the average delay.
\end{theorem}
\begin{remark}\normalfont
    Compared to the existing $\mathcal{O}(nd^{1/3}T^{2/3})$ regret bound,  DBGD-NF reduces the effect of delay and  achieves a better $\mathcal{O}(n\bar{d}^{1/3}T^{2/3})$ regret bound. It is worth noting that DBGD-NF requires the prior knowledge of the average delay to set the learning rate. \citet{quanrud2015online} also encounter this issue and  introduce a simple solution by utilizing the doubling trick \cite{cesa1997use} to adaptively adjust the learning rate, thereby overcoming this limitation, which we can also employ  to attain an equivalent $\mathcal{O}\left(n\Bar{d}^{1/3}T^{2/3}\right)$ bound. 
\end{remark}

\begin{algorithm}[!t]
\caption{DOGD-NF}
	\label{alg:0}
	\begin{algorithmic}[1]
     \REQUIRE Learning rate $\eta$
     \STATE Initialize $\x_1 \in [0,1]^n$
     \FOR{$t=1$ to $T$}
     \STATE Let $1 \geq \x_{t,\pi_{(1)}} \geq \x_{t,\pi_{(2)}} \geq \dots \x_{t,\pi_{(n)}} \geq 0 $ be the sorted entries in decreasing order with $A_{t,i} =\{\pi_{(1)}, \dots, \pi_{(i)}\}$ for all $i \in [n]$ and $A_{t,0} = \emptyset$. Define $\x_{t,\pi_{(0)}}=1, \x_{t,\pi_{(n+1)}}=0$
     \STATE For $0 \leq i \leq n$, calculate $\lambda_{t,i} = \x_{t,\pi_{(i)}}-\x_{t,\pi_{(i+1)}}$
     \STATE Sample $S_t$ from the distribution defined in (\ref{s_t1})
     \STATE Observe the loss $f_t(S_t)$ 
     \STATE Compute the estimated gradient $\g_t$ by (\ref{gt1}) and incur a delay $d_t \geq 1$
     \STATE Receive the gradient set $\{\g_k|k\in\F_t\}$
     \STATE Update  $\x_{t}$ according to (\ref{update1})
     \ENDFOR
	\end{algorithmic}
\end{algorithm}

Additionally, we also observe that  DBGD-NF can be slightly adjusted to the full-information setting to derive a regret bound that relies on the average delay. Our modified method, DOGD-NF,  is summarized in Algorithm \ref{alg:0}.  In each round $t$, we sample $S_t$  from the distribution 
\begin{equation}\label{s_t1}
    P(S_t = A_{t,i}) = \lambda_{t,i},
\end{equation}
where $A_{t,i} =\{\pi_{(1)}, \dots, \pi_{(i)}\}, A_{t,0} = \emptyset$ and $\lambda_{t,i} = \x_{t,\pi_{(i)}}-\x_{t,\pi_{(i+1)}}.$ Similar to DOAGD \cite{lin2022online}, we also employ the convex relaxation based on the Lovász extension to compute the approximate subgradient
\begin{equation}\label{gt1}
    \g_{t,\pi(i)}=f_{t}(A_{t,i})-f_{t}(A_{t,i-1}).
\end{equation}
Finally, we use all the available gradients to perform a gradient descent step
\begin{equation}\label{update1}
    \x_{t+1} = \Pi_{[0,1]^n} \left[ \x_t - \sum_{k \in \F_t} \g_k \right].
\end{equation}

Then we present the theoretical guarantee of Algorithm \ref{alg:0}.
\begin{theorem}\label{thm:0}
    Under Assumption \ref{ass:1} and Assumption \ref{ass:2}, by setting $\eta = \frac{\sqrt{n}}{L\sqrt{\Bar{d}T}}$, DOGD-NF ensures
    \begin{equation*}
        \mathbb{E}\left[\textbf{Reg}_{\alpha,\beta}(T)\right] \leq \mathcal{O}(\sqrt{n\bar{d}T}).
    \end{equation*}
\end{theorem}

To better showcase the improvement of our results, we provide an example that clearly demonstrates the enhancement through the better exponents of $T$.

\begin{example}
    Consider a situation:  $d_{1:T}$ satisfy $d_{1:\sqrt{T}} = \frac{T}{2}$ and $d_{\sqrt{T}+1:T} = 1$. Our methods achieve $\mathcal{O}(\sqrt{n}T^{3/4})$ and $\mathcal{O}(nT^{5/6})$ regret bounds, while the bounds of DOAGD and DBAGD are $\mathcal{O}(\sqrt{n}T)$ and $\mathcal{O}(nT)$, respectively.  
\end{example}

\subsection{Decoupled Result}
% Since BOAGD \cite{lin2022online} employs the one-point gradient estimator, the bound of  its gradient is $\mathcal{O}(n/\sqrt{\mu})$. Due to the delayed feedback, it  suffers an additional $\mathcal{O}(ndT\eta/\sqrt{\mu})$ regret over the rounds $T$, where we set $\eta_t = \eta$ just for simplicity of the discussion. This is the primary reason why its regret bound is the product of the delayed effect and the $\mathcal{O}(nT^{2/3})$ \cite{lin2022online} bound in the  bandit setting without delayed feedback.  

While BOGD-NF achieves a SOTA regret bound with respect to the average delay $\bar{d}$, its dependence on the maximum delay $d$ remains suboptimal. In the following, we introduce a modification of BOGD-NF aimed at improving the regret dependence on the maximum delay. Inspired by \citet{wan2024improved}, we combine DBGD-NF  with a blocking update technique~\cite{zhang2019online,garber2020improved,AAAI:2023:Wang,AAAI:2024:Wang} to  reduce the effect of delays. The essential idea is dividing total $T$ rounds into blocks with size $K$ and choosing the decision $S_t$ from the same mixture distribution per block, so that we can reduce the bound of the estimated gradients suffered within each block.  It is not hard to verify that the bound on the sum of gradients within a block is 
\begin{equation*}
 \E\left[\Norm{\sum_{t=mK+1}^{(m+1)K} \hat{\g}_t}\right] \leq \mathcal{O}\left(n\left(\sqrt{K/\mu}+K\right)\right).   
\end{equation*}
By choosing a proper block size $K  = \Theta(1/\mu)$, this bound can be enhanced to $\mathcal{O}\left(nK\right)$, which is smaller than the $\mathcal{O}(nK/\sqrt{\mu})$ bound of BDAGD. Based on this blocking update mechanism, we reduce the effect of delays on the regret from $\mathcal{O}(n\eta d T\sqrt{\mu})$ to $\mathcal{O}(n\eta d T)$, so that we decouple the effect of the delays and gradient estimator. 
\begin{algorithm}[!t]
\caption{BDBGD-NF}
	\label{alg:1}
	\begin{algorithmic}[1]
     \REQUIRE $\eta, \mu \in (0,1), K$
     \STATE Initialize point $\x_1 \in [0,1]^n$, set $\y_1 = \x_1$ and set the gradient pool for each block $\mathcal{P}_i = \emptyset, i = 1,...,\lceil T/K\rceil$
     \FOR{$m=1$ to $\lceil T/K\rceil$}
     \STATE Set block pool $\mathcal{A}_m = \emptyset$
     \FOR{time step $t=(m-1)K+1$ to $\min\{mK,T\}$}
     \STATE Choose $\x_t = \y_m$
     \STATE Let $1 \geq \x_{t,\pi_{(1)}} \geq \x_{t,\pi_{(2)}} \geq \dots \x_{t,\pi_{(n)}} \geq 0 $ be the sorted entries in decreasing order with $A_{t,i} =\{\pi_{(1)}, \dots, \pi_{(i)}\}$ for all $i \in [n]$ and $A_{t,0} = \emptyset$. Define $\x_{t,\pi_{(0)}}=1, \x_{t,\pi_{(n+1)}}=0$
     \STATE For $0 \leq i \leq n$, calculate $\lambda_{t,i} = \x_{t,\pi_{(i)}}-\x_{t,\pi_{(i+1)}} $ 
     \STATE Sample $S_t$ from $P \left(S_{t}=A_{t,i}\right) = (1-\mu)\lambda_{t,i} + \frac{\mu}{n+1}$
     \STATE Observe the loss $f_t(S_t)$ 
     \STATE Calculate $\hat{f}_{t,i}=\frac{\mathbf{1}\left(S_{t}=A_{t,i}\right)}{\left(1-\mu\right) \lambda_{t,i}+\frac{\mu}{n+1}} f_{t}\left(S_{t}\right)$
     \STATE Compute the estimated gradient  $\hat{\g}_{t,\pi(i)}=\hat{f}_{t,i}-\hat{f}_{t,i-1}$ and incur a delay $d_t \geq 1$
     \STATE Receive the set $\{\hat{\g}_k|k\in\F_t\}$ and update each gradient to its gradient pool $\mathcal{P}_j =\mathcal{P}_j \bigcup \{\hat{\g}_k\}$,  where $j = \lceil k/K \rceil$ is the block that $\hat{\g}_k$ belongs to
     \ENDFOR
     \STATE If $|\mathcal{P}_i|=K$, $\mathcal{A}_m = \mathcal{A}_m \bigcup \{i\}$, which denotes the index of the block where all the queries arrive. 
     \STATE Perform gradient descent update to   $\y_{m+1} = \Pi_{[0,1]^{n}}\left[\y_m - \sum_{i\in \mathcal{A}_m}\sum_{\hat{\g}_k\in \mathcal{P}_i} \hat{\g}_k \right]$
     \STATE For $i\in \mathcal{A}_m$, set $\mathcal{P}_i = \emptyset$
     \ENDFOR
	\end{algorithmic}
\end{algorithm}
\begin{figure*}[ht]
	\begin{center}
		\subfigure[$d=10$]{\includegraphics[width=0.3\textwidth]{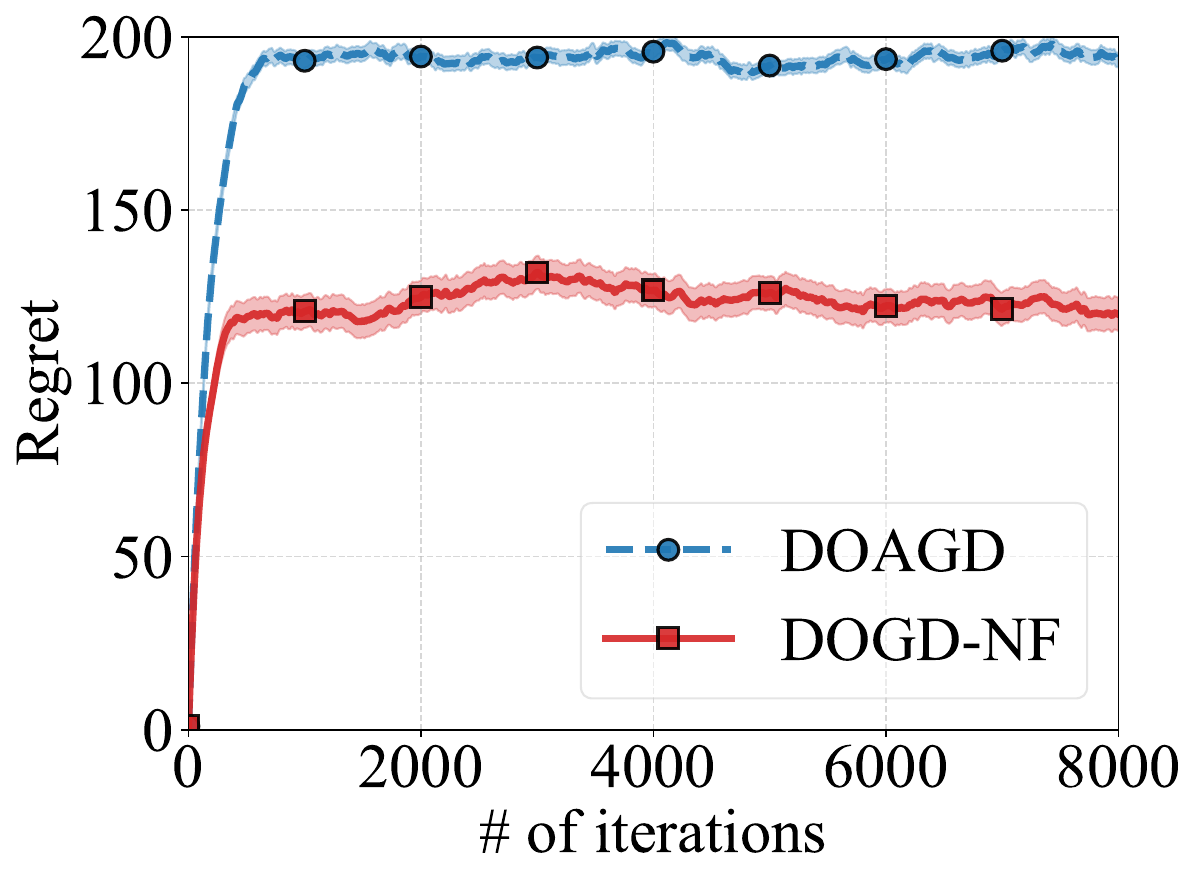}}
   \subfigure[$d=20$]{\includegraphics[width=0.3\textwidth]{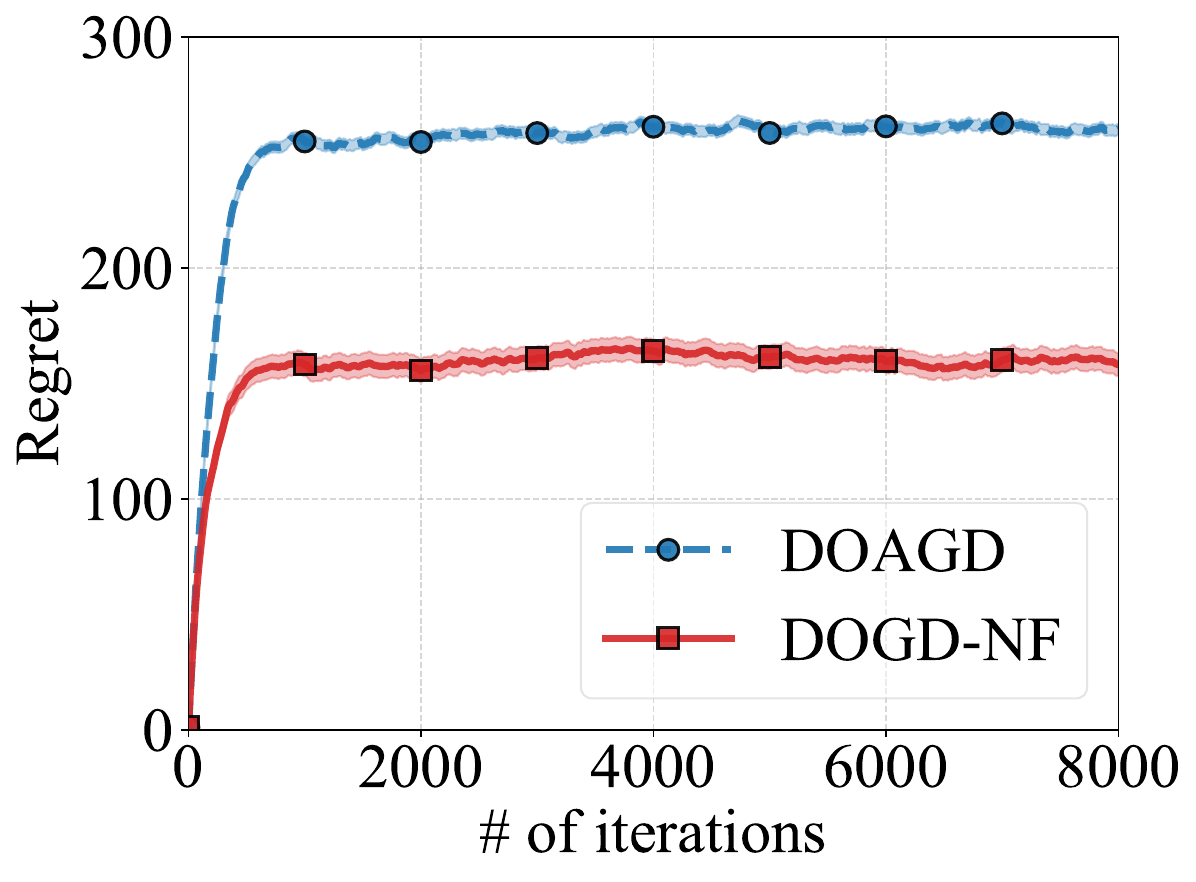}}
   \subfigure[$d=500$]{\includegraphics[width=0.3\textwidth]{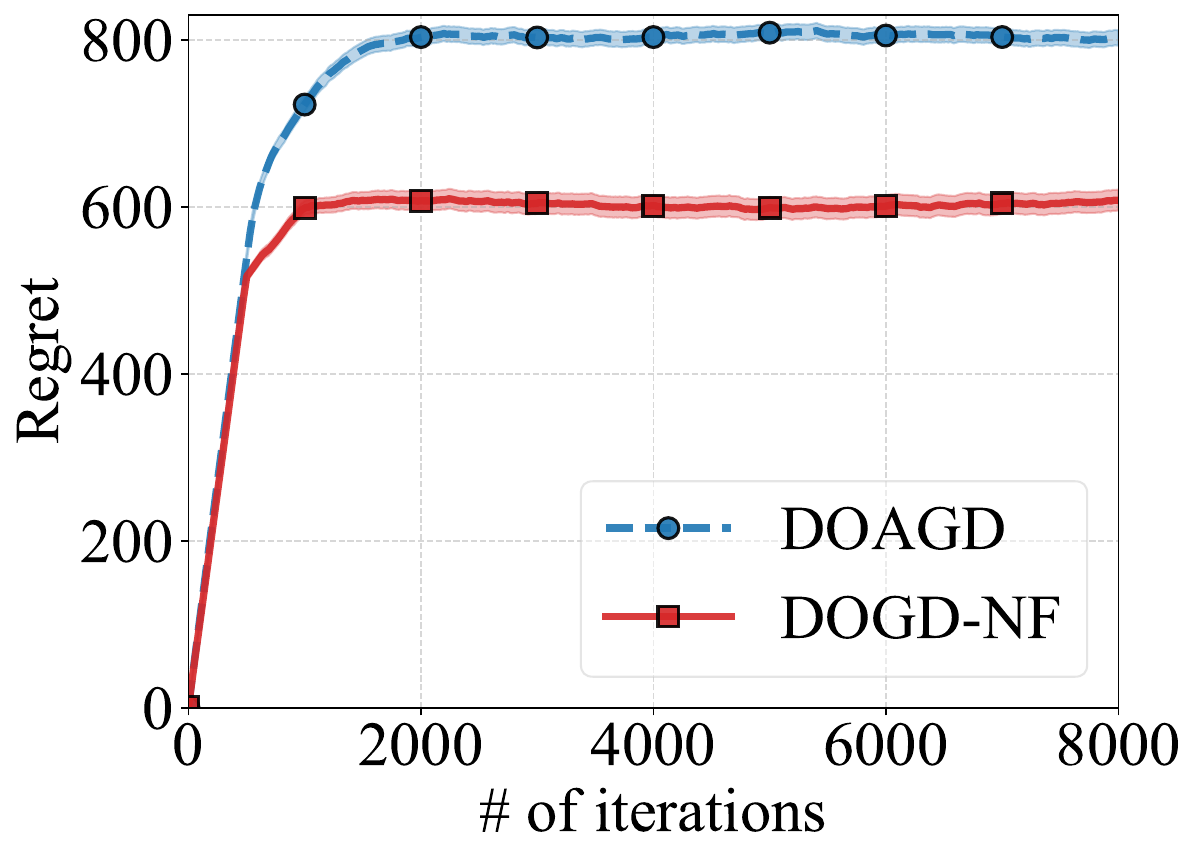}}
		\caption{Results for full-information.}
		\label{fig:0}
	\end{center}
\end{figure*}
\begin{figure*}[t]
	\begin{center}
		\subfigure[$d=10$]{\includegraphics[width=0.3\textwidth]{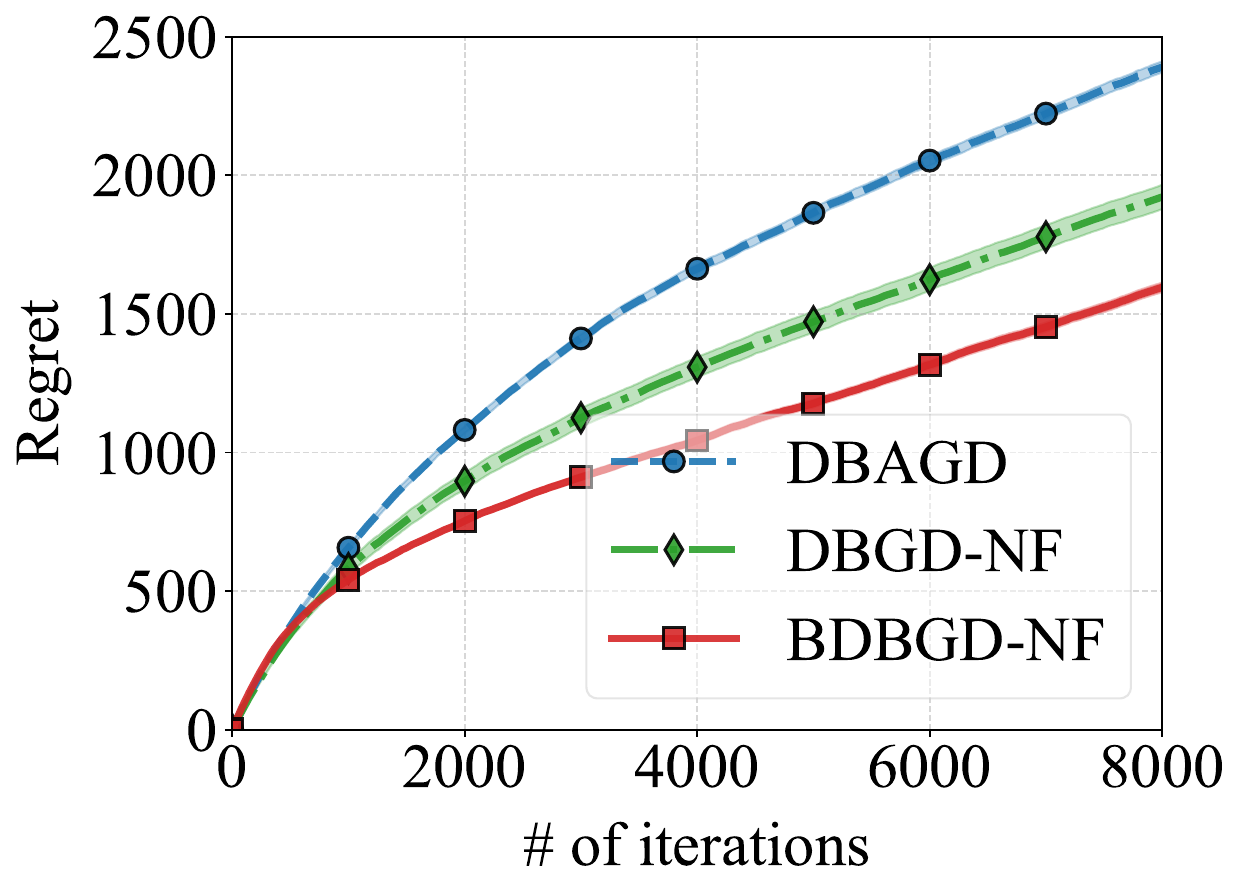}}
   \subfigure[$d=20$]{\includegraphics[width=0.3\textwidth]{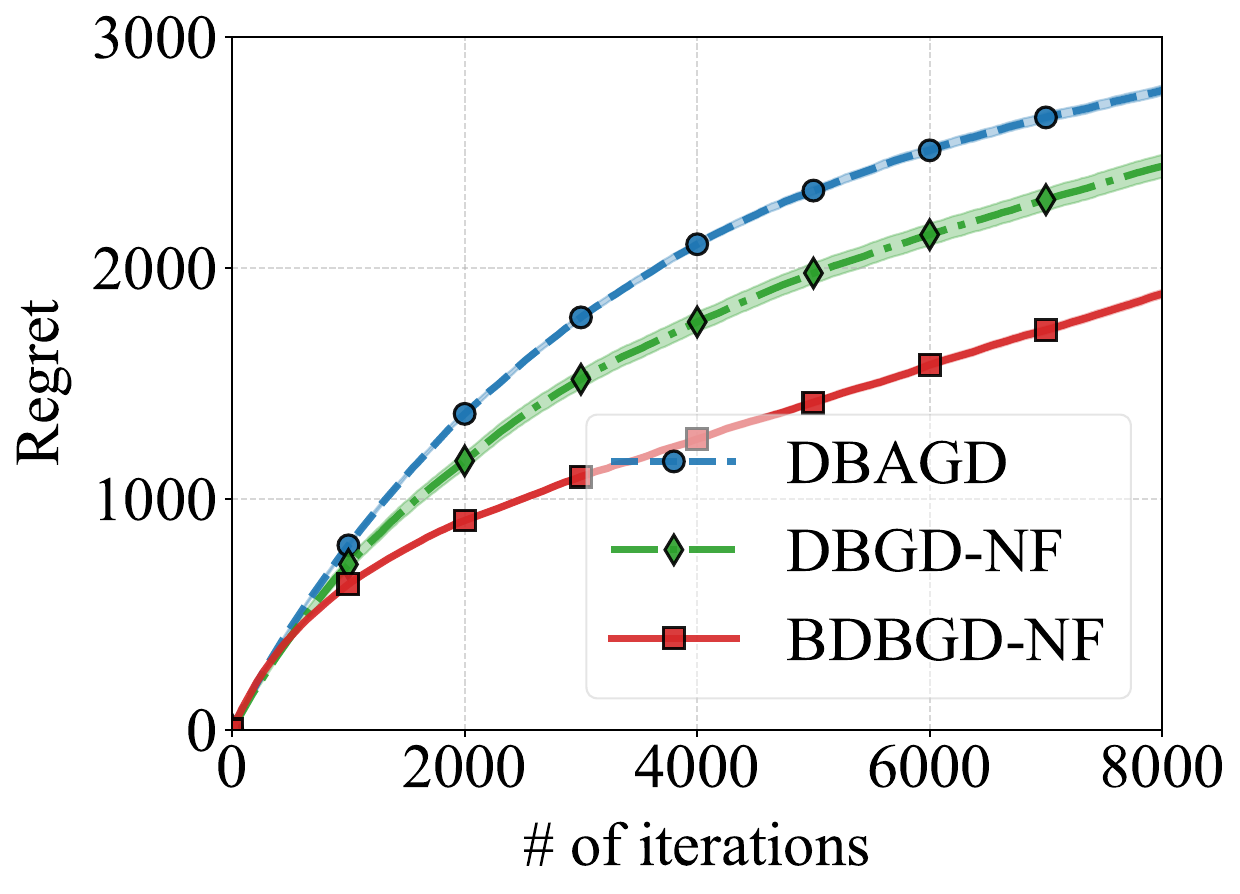}}
   \subfigure[$d=500$]{\includegraphics[width=0.3\textwidth]{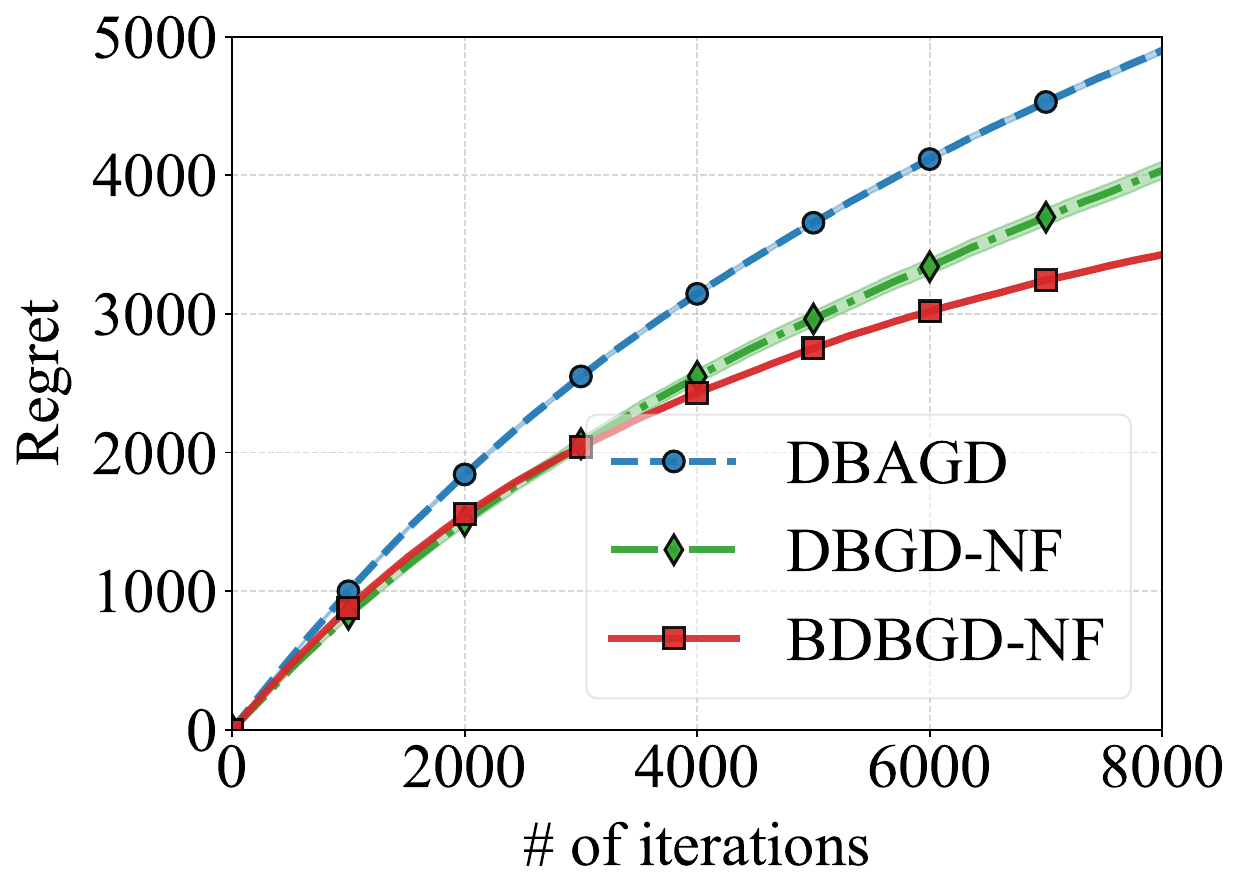}}
		\caption{Results for bandit feedback.}
		\label{fig:1}
	\end{center}
\end{figure*}

Our designed algorithm, BDBGD-NF, is presented in Algorithm \ref{alg:1}.  We first simply initialize $\x_1, \y_1$ to be any point in $[0,1]^n$ and available gradient set for the each block  $\mathcal{P}_i = \emptyset$. In each block $m$, we sample the decision set $S_t$ from the same mixture distribution (\ref{distribution}) related to $\y_m$ and employ the one-point gradient estimator to derive the gradient.

 At the end of each block $m$, we will identify the block where all the queries are obtained and use the all gradients from available blocks to update $\y_m$
\begin{equation}\label{update}
     \y_{m+1} = \Pi_{[0,1]^{n}}\left[\y_m - \sum_{i\in \mathcal{A}_m}\sum_{\hat{\g}_k\in \mathcal{P}_i} \hat{\g}_k \right].
\end{equation}

% Remarkably, compared to \citet{wan2024improved}, BDBGD-NF is specifically designed for online nonsubmodular optimization. A key challenge in designing BDBGD-NF is that if we perform gradient descent on available gradients individually at the end of each block, we cannot use  reduce the variance of the gradient estimator. To address this, we update the decision only using the gradients from blocks where all information are available, which has not been used in previous work.

In the following, we  present the following theorem to establish the  theoretical guarantee for BDBGD-NF.

\begin{theorem}\label{thm:1}
Under Assumption \ref{ass:1} and Assumption \ref{ass:2}, by setting $K = T^{\frac{1}{3}}, \mu = nT^{-1/3}, \eta = \min\{\frac{1}{LT^{2/3}}, \frac{1}{L\sqrt{dT}}\}$, BDBGD-NF ensures 
\begin{align*}
    \E\left[\textbf{Reg}_{\alpha,\beta}(T)\right] \leq \mathcal{O}\left(n(T^{\frac{2}{3}} + \sqrt{dT})\right).
\end{align*}
\end{theorem}
\begin{remark}\normalfont
When $d = \mathcal{O}(T^{1/3})$, the regret bound of our method matches the previous $\mathcal{O}(nT^{2/3})$ regret bound \cite{lin2022online} in  the  bandit setting without delayed feedback. Otherwise, this regret bound is also on the same order in terms of $d$ and $T$ with the  $\mathcal{O}(\sqrt{n\Bar{d}T})$ regret in the full-information  setting with delayed feedback in the worst case, where $\Bar{d} = \Theta(d)$. Moreover, it is better than the former $\mathcal{O}\left(n\Bar{d}^{1/3}T^{2/3}\right)$ bound when $d = o (\bar{d}^{2/3}T^{1/3})$. 
\end{remark}
% \begin{remark}\normalfont
%     One might notice that our BDBGD-NF may maintain $\lceil T/K\rceil$ gradient pools, potentially resulting in high space complexity. However, as the query is delayed by at most $d$ rounds, there are only at most $\lceil d/K\rceil+1$ gradient pools at any time.
% \end{remark}

\section{Experiments}

In this section, we evaluate the effectiveness of our proposed methods through numerical experiments. We compare our methods with the SOTA methods, DOAGD and  DBAGD~\cite{lin2022online}  on structured sparse learning with delayed feedback. When it comes to hyper-parameter tuning,  we set $\eta = \frac{\sqrt{n}}{L\sqrt{\Bar{d}T}}$ for DOGD-NF, $\eta = \frac{1}{L\Bar{d}^{1/3}T^{2/3}}$ and $\mu = \frac{q n \Bar{d}^{1/3}}{T^{1/3}}$ for DBGD-NF and  $\eta = \min\{\frac{1}{LT^{2/3}}, \frac{1}{L\sqrt{dT}}\}$, $\mu = \frac{q n}{T^{1/3}}$ and $K = T^{1/3}$ for BDBGD-NF. As for other methods, we choose $\eta = \frac{\sqrt{n}}{L\sqrt{dT}}$  for DOAGD,  and  $\eta = \frac{1}{Ld^{1/3}T^{2/3}}$ and $\mu = \frac{q n d^{1/3}}{T^{1/3}}$ for DBAGD, which are according to Theorem 5.2 and Theorem 5.4 in \citet{lin2022online}. Moreover, we perform a  grid search to select the   parameter  $q$  from the set $\{0.01,0.1,1\}$. All the experiments are conducted in Python 3.7 with two $3.1$ GHz Intel  Xeon Gold 6346 CPUs and 32GB memory.

\textbf{Setup.}  We conduct experiments on structured sparse learning defined in (\ref{exper}).   Following the setup in \citet{lin2022online}, we choose $F_t(S) = \max(S) - \min(S) + 1$ for all $S \neq \emptyset$ and $F(\emptyset)=0$. We consider a simple linear regression problem, where the  sparse optimal solution $\x^* \in \mathbb{R}^n$ only consists  $k$ consecutive $1$, with the remaining positions being $0$. We compute $\y_t = A_t\x^* + \mathbf{\epsilon}_t$, where each row of $A_t \in \mathbb{R}^{s \times n}$ is a vector i.i.d. sampled from a Gaussian distribution and $\mathbf{\epsilon}_t \in \mathbb{R}^{s}$ is a Gaussian noise vector with standard deviation $0.1$. We choose the square loss $\ell_t(\x) = \Norm{A_t\x - \y_t}^2/2$ and random delays in our experiments, i.e., $1 \leq d_t \leq d$, where $d$ is the max delay. For all experiments, we set rounds $T = 8000$ such that the block $K$  is $20$, dimension $n = 10$, number of samples $s=128$, trade-off parameter $\gamma = 0.1$ and sparse parameter $k = 2$. 

 \textbf{Results.} We report the regret against the number of rounds in full-information and bandit settings under different maximum delays $d$ within the set $\{10,20,500\}$ in Figure \ref{fig:0} and Figure \ref{fig:1}, respectively. For the delays $d_t$, we sample them uniformly at random from the range $[1,d]$. We observe that DOGD-NF suffers less loss compared to DOAGD under full-information with delayed feedback because it utilizes all available gradients rather than just the oldest one. When the maximum delay increases, the superiority becomes more pronounced due to the large gap between the average delay $\Bar{d}$ and maximum delay $d$. As evident from Figure \ref{fig:1}, our DBGD-NF also experiences less loss, which is consistent with our theories. Additionally, BDBGD-NF obtains the best performance and  consistently yields lowest regret as the delay changes in the  bandit setting, which aligns with its theoretical guarantee.

\section{Conclusion and Future Work}

In this paper, we study the online nonsubmodular optimization with delayed feedback in the bandit setting and develop several algorithms to improve the existing regret bound. Firstly, our BDGD-NF and DOGD-NF achieve better \(\mathcal{O}(n\Bar{d}^{1/3}T^{2/3})\) and \(\mathcal{O}(\sqrt{n\Bar{d}T})\)  regret bounds for bandit and  full-information settings, respectively, which are relevant to the average delay. Furthermore,  to decouple the joint effect of delays and bandit feedback, we combine BDGD-NF with a blocking update technique. Our  BDBGD-NF obtains a superior  $\mathcal{O}(n(T^{2/3} + \sqrt{dT}))$ regret bound. Finally, the experimental results also demonstrate the effectiveness of our methods.

One might notice that the regret bound of BDBGD-NF depends on the maximum delay rather than the average delay, unlike the former  results. In the future, we will investigate how to develop a decoupled algorithm to obtain the regret bound  that relies on the average delay. Moreover, we will try to deal with the online nonsubmodular optimization in the non-stationary environments.

\section{Acknowledgments}
This work was partially supported by NSFC (U23A20382, 62361146852), the Collaborative Innovation Center of Novel Software Technology and Industrialization, and the Open Research Fund of the State Key Laboratory of Blockchain and Data Security, Zhejiang University. The authors would like to thank the anonymous reviewers for their constructive suggestions.

\bibliography{aaai25}

\end{document}